\documentclass[11pt,letterpaper]{article}
\usepackage[T1]{fontenc}
\usepackage{lmodern}
\usepackage[margin=1in]{geometry}
\usepackage[protrusion=false]{microtype}
\usepackage{titling}
\pretitle{\begin{center}\Large\bfseries}
\posttitle{\par\end{center}\vskip 0.5em}
\predate{}
\postdate{}
\usepackage[hyphens]{url}
\usepackage{graphicx}
\graphicspath{{figures/}}
\usepackage[round]{natbib}

\usepackage[font=small,labelfont=bf]{caption}
\usepackage{placeins}
\usepackage{booktabs,multirow,array}
\usepackage{algorithm,algorithmic}
\usepackage{amssymb,amsthm,bm,mathtools}
\usepackage{enumitem}
\setlist{itemsep=3pt,topsep=6pt,parsep=0pt}
\usepackage[hidelinks]{hyperref}
\hypersetup{
 pdftitle={Neural Supervised Clustering for Learning Individualized Treatment Rules with Many Treatments},
 pdfauthor={Daixi Huang; Yixin Yuan; Yifan Dai; Yufeng Liu},
 pdfsubject={Treatment-rule learning with supervised neural treatment clustering}
}
\makeatletter
\let\inputtablerows\@@input
\makeatother

\theoremstyle{definition}

\theoremstyle{remark}

\newcommand{\E}{\mathbb{E}}

\newcommand{\argmax}{\operatorname*{arg\,max}}
\newcommand{\argmin}{\operatorname*{arg\,min}}

\title{Neural Supervised Clustering for Learning\\
Individualized Treatment Rules with Many Treatments}
\author{%
Daixi Huang\textsuperscript{1} \quad
Yixin Yuan\textsuperscript{2} \quad
Yifan Dai\textsuperscript{3} \quad
Yufeng Liu\textsuperscript{1}\thanks{Corresponding author: \href{mailto:yufliu@umich.edu}{\texttt{yufliu@umich.edu}}.}\\[8pt]
\small\textsuperscript{1}University of Michigan
\qquad\textsuperscript{2}Yale University\\[2pt]
\small\textsuperscript{3}University of North Carolina at Chapel Hill\\[7pt]
\small\begin{tabular}{@{}l@{\qquad}l@{}}
\href{mailto:daixih@umich.edu}{\texttt{daixih@umich.edu}} &
\href{mailto:amelia.yuan@yale.edu}{\texttt{amelia.yuan@yale.edu}}\\[2pt]
\href{mailto:yfd@unc.edu}{\texttt{yfd@unc.edu}} &
\href{mailto:yufliu@umich.edu}{\texttt{yufliu@umich.edu}}
\end{tabular}}
\date{}

\begin{document}
\maketitle
\begin{abstract}
Learning an individualized treatment rule (ITR) with many treatments requires estimating treatment-specific effects from limited observations per arm. Treatments with similar conditional responses can share information, but linear contrast models may miss nonlinear heterogeneity. We propose nonlinear supervised clustering fusion (NSCF), a neural Q-learning framework with a shared nonlinear covariate representation and low-dimensional treatment embeddings. Pairwise fusion groups treatments with similar learned contrasts, yielding an ITR and an interpretable partition. An alternating algorithm combines neural training, embedding fusion, validation-based partition selection, and unpenalized grouped refitting. NSCF lowers regret relative to a strong neural reference in nonlinear exact-sharing simulations; perturbations characterize the limits of hard sharing. In a melanoma xenograft application, NSCF improves average policy value over the neural reference. These results support nonlinear representation learning with supervised treatment fusion when treatment effects exhibit shared structure.

\end{abstract}

\section{Introduction}

Data-driven individualized decision-making has become increasingly important in precision medicine \citep{collins2015precision,kosorok2019precision}. Because treatment effects can vary substantially across patients, assigning the same treatment to every individual is often suboptimal. An individualized treatment rule (ITR) instead aims to recommend the treatment that maximizes the expected clinical outcome based on patient-specific covariates, such as demographic characteristics, clinical measurements, and molecular biomarkers.

A large body of work has focused on estimating optimal ITRs. Regression- or contrast-based approaches, including Q-learning and A-learning, estimate conditional outcome or treatment-contrast functions and then recommend the treatment with the largest predicted benefit \citep{murphy2003optimal,qian2011performance,chakraborty2013statistical,wallace2015doubly,shi2018high}. Policy-based approaches instead optimize empirical or surrogate objectives tied to policy value, including outcome-weighted learning, residual weighted learning, empirical welfare maximization, and related robust policy-learning formulations \citep{zhang2013robust,zhao2012estimating,zhao2015new,zhou2017residual,kitagawa2018who,kallus2018balanced,athey2021policy}. For multi-arm settings, direct-learning and margin-based formulations further address the multicategory decision structure \citep{qi2020multiarmed,zhang2020multicategory}. These methods have achieved considerable success when the number of treatment options is relatively small.

Modern precision medicine increasingly involves many candidate treatments but limited observations per treatment. Estimating treatment-specific contrasts can then be statistically inefficient and yield unstable recommendations \citep{qi2020multiarmed,zhang2020multicategory}. When some treatments have similar response patterns, borrowing information across them offers a way to improve estimation and obtain interpretable recommendations. Grouping provides a structural restriction that reduces the number of distinct treatment-response functions to be estimated.

Supervised clustering methods exploit latent grouping structure by combining a predictive loss with a fusion penalty that encourages similar supervised parameters to be estimated together \citep{wei2013latent,shen2010grouping,ke2015homogeneity,ma2017concave,chen2021identifying}. For many-treatment ITRs, fusion-based approaches estimate treatment partitions through penalized contrast or working models \citep{ma2022many,zhu2025doubly}. Supervised clustering with adaptive fusion (SCAF), one of our comparison methods, applies adaptive fusion to treatment-specific linear contrast vectors.

Treatment grouping has also been studied beyond linear contrast models. Group Outcome Weighted Learning (GROWL) jointly learns treatment partitions and group-based decision rules, allowing nonlinear decision functions in reproducing kernel Hilbert spaces \citep{ma2023growl}. The calibration-weighted fusion method of \citet{zhu2025doubly} addresses covariate imbalance through a penalized working model and permits flexible policy learning in the fused treatment space. Meanwhile, neural ITR methods provide flexible models for nonlinear treatment heterogeneity \citep{mi2019bagging,xu2024optimal}. Although a shared neural representation enables information sharing across treatments, it does not by itself produce an explicit treatment partition. This motivates coupling neural outcome modeling with supervised fusion to learn both nonlinear treatment effects and interpretable treatment groups.

We develop nonlinear supervised clustering fusion (NSCF), a neural Q-learning framework for many-treatment ITR estimation. NSCF couples treatment-specific embeddings with a shared nonlinear representation of patient covariates. Starting from a pretrained neural outcome model, it alternates between fusion with the representation held fixed and unpenalized grouped refitting that updates the representation. Validation data guide model and partition selection. Within the fitted model, fused embeddings induce identical treatment-contrast functions across covariate values. Thus, treatments with similar learned response patterns share information within a nonlinear outcome model.

\paragraph{Contributions.}
Our main contributions are as follows.
\begin{enumerate}
    \item We propose NSCF, which combines neural Q-function representation learning with supervised fusion of treatment-specific embeddings. The resulting treatment partition imposes explicit sharing of learned contrast functions within a flexible nonlinear outcome model.

    \item We formulate a convex conditional fusion problem for a fixed representation and develop a validation-guided alternating procedure with grouped refitting. This separates fusion within a common representation from subsequent representation updates.

    \item We evaluate NSCF against neural and classical baselines across sample sizes, treatment-group structures, and departures from exact sharing. Component ablations examine representation updates and grouped refitting, and a melanoma PDX application assesses individualized recommendations alongside a training-selected constant strategy.
\end{enumerate}

Section~\ref{sec:setup} presents the method. Sections~\ref{sec:experiments}--\ref{sec:realdata} report simulations and the PDX application, and Section~\ref{sec:discussion} discusses the findings and limitations. Additional details are provided in the appendix.

\section{Methodology}
\label{sec:setup}

\subsection{Single-stage Q-learning and treatment sharing}
Consider independent observations $\mathcal D=\{(X_i,A_i,Y_i)\}_{i=1}^n$, where $X_i\in\mathbb R^p$, $A_i\in\{1,\ldots,K\}$, and larger outcomes are preferable. An individualized treatment rule (ITR) $d$ has value $\mathcal V(d)=\E\{Y(d(X))\}$. Under consistency, conditional exchangeability $\{Y(a)\}_{a=1}^K\perp A\mid X$, and positivity, the conditional potential-outcome mean is identified by $Q(x,a)=\E(Y\mid X=x,A=a)$ \citep{rubin1974estimating}. Hence $d^\star(x)\in\argmax_a Q(x,a)$ maximizes value. These assumptions concern identification; they do not follow from fitting a neural network.

We use single-stage, regression-based Q-learning \citep{murphy2003optimal,qian2011performance}, rather than a sequential reinforcement-learning procedure. A fitted outcome model induces $\widehat d(x)\in\argmax_a\widehat Q(x,a)$. For integrable conditional means and estimation errors,
\begin{equation}
 \mathcal V(d^\star)-\mathcal V(\widehat d)
 \leq 2\E\max_a|\widehat Q(X,a)-Q(X,a)|.
 \label{eq:regret-bound}
\end{equation}
%Appendix~\ref{app:theory} derives Eq.~\eqref{eq:regret-bound}.
The bound concerns prediction errors across treatments, whereas factual
MSE measures prediction error at the observed treatment.
We therefore evaluate both predictive accuracy and policy performance.

When $K$ is large relative to $n$, separate arm-specific fits use little information per treatment. We instead allow treatments to share an outcome-trained representation and, when supported by validation, a contrast function. Write
\begin{equation}
 Q_{\theta,\Gamma}(x,a)=m_\theta(x)+h_\theta(x)^\top\gamma_a,
 \qquad \Gamma\in\mathbb R^{K\times r},
 \label{eq:neural-q}
\end{equation}
where $h_\theta(x)\in\mathbb R^r$ is a shared ReLU network and $m_\theta(x)=b+h_\theta(x)^\top\beta$ is the prognostic head. The rows of $\Gamma$ are treatment embeddings. Thus a single network jointly estimates the main effect and treatment contrasts; the architecture is not itself a new general-purpose neural architecture. A one- or two-hidden-layer network is used, with independent $N(0,0.02^2)$ initial treatment coordinates and dropout during training. %Widths and training rules are given in Appendix~\ref{app:implementation}.

The estimated decision depends only on $h_{\widehat\theta}(x)^\top\widehat\gamma_a$, because the prognostic term cancels across arms. If $\gamma_a=\gamma_b$, the two fitted contrasts coincide for every $x$. This gives equality of embeddings a functional meaning without interpreting their individual coordinates. The converse need not hold if the representation is redundant. Grouping is \emph{supervised}: outcomes determine the representation and embeddings, rather than an unsupervised distance between treatment labels.

\subsection{Conditional fusion and grouped refitting}
For fixed $\theta$, define
\begin{align}
 f_\theta(\Gamma)&=\frac1n\sum_{i=1}^n\mathcal L\{Y_i,Q_{\theta,\Gamma}(X_i,A_i)\},\nonumber\\
 F_{\theta,\lambda}(\Gamma)&=f_\theta(\Gamma)+\lambda\sum_{a<b}\|\gamma_a-\gamma_b\|_1.
 \label{eq:fusion-objective}
\end{align}
Here $\mathcal L$ is a convex regression loss; simulations use Huber loss \citep{huber1964robust}. Pairwise coordinatewise $\ell_1$ fusion encourages shared embedding coordinates \citep{tibshirani2005fused}; a treatment group requires equality of the full rows. We use the unnormalized sum over all pairs and coordinates. Accordingly, the numerical meaning of $\lambda$ depends on embedding scale, dimension, and the number of treatments.

The coordinatewise $\ell_1$ penalty admits an auxiliary proximal update
that separates across embedding coordinates. Each coordinate subproblem
is solved exactly by isotonic regression before numerical snapping,
as detailed below. The group count is selected along the fusion path
using validation data. In a controlled fixed-representation comparison,
$\ell_1$ and group-$\ell_2$ fusion selected the same partition in
59 of 60 neural-feature cases and yielded similar post-refit regret. Because fusion acts on embedding coordinates,
its path depends on the representation's parameterization.

Penalized embeddings are used to propose partitions. Given a partition $g:\{1,\ldots,K\}\to\{1,\ldots,G\}$, we refit
\[
 Q^{\rm refit}(x,a)=m_{\theta'}(x)+h_{\theta'}(x)^\top\zeta_{g(a)}
\]
without the fusion penalty but retaining weight decay; this is what we mean by ``unpenalized'' refitting. All network parameters may update, with one embedding per group. Distinct partitions are refitted once per cycle from the cycle-start weights, averaging treatment embeddings within each group. This separates partition discovery from estimation under the resulting equality constraints.

\subsection{Validation-guided alternating algorithm}
NSCF starts at a strongly trained unfused Neural Q model and alternates conditional fusion with unpenalized grouped refitting (Algorithm~\ref{alg:nscf}). Each conditional path uses an exact coordinatewise auxiliary proximal calculation before numerical snapping. The algorithm returns the validation-selected model from these cycles. Grouped refitting is performed \emph{within} each cycle; no additional post-selection restart is required.

\begin{algorithm}[!t]
\caption{NSCF: fusion discovery and refitting}
\label{alg:nscf}
\small
\begin{algorithmic}[1]
\STATE \textbf{Input:} training/validation data, grid $\Lambda$,
cycle limit $C$, fixed training/selection rules.
\STATE \textbf{Initialize:} Train an unfused Neural Q anchor
with validation MSE $E_0$; initialize the working model.
\FOR{$c=1,\ldots,C$, with early stopping}
 \STATE Save the cycle-start weights; freeze the representation
 and prognostic head for fusion.
 \FOR{$\lambda\in\Lambda$ in increasing order}
  \STATE \textbf{Fuse:} Warm-start $(\Gamma,Z,U)$ and apply
  the finite-budget updates in Eq.~\eqref{eq:admm-updates}.
  \STATE Extract groups from rounded, tolerance-matched $Z$ rows.
  \STATE \textbf{Refit:} Fit each distinct full grouped network from
  cycle-start weights without the fusion penalty; evaluate validation MSE.
 \ENDFOR
 \STATE \textbf{Select:} Among refits with MSE $\leq E_0$,
 choose the fewest groups within the prescribed relative
 tolerance of the best eligible MSE. If none is eligible,
 use the minimum-MSE refit.
 \STATE Retain its partition and validation MSE $E_c$;
 use the selected model for the next cycle.
 \STATE Record $S_c=E_c+\omega\Delta_c$, where
 $\Delta_c=\|Z_c^*-Z_{c-1}^*\|_F$ and $\Delta_1=0$.
\ENDFOR
\STATE \textbf{Finalize:} Compare the score-based and near-best,
fewest-group cycle candidates by validation MSE.
%(Appendix~\ref{app:selection}).
\STATE \textbf{Return} the selected model $\widehat Q$, partition,
and $\widehat d(x)\in\argmax_a\widehat Q(x,a)$.
\end{algorithmic}
\end{algorithm}

For the conditional path, introduce $Z=\Gamma$ and a scaled dual variable $U$. One inner iteration is
\begin{align}
 \Gamma^{q+1}&\approx\argmin_\Gamma f_\theta(\Gamma)+\tfrac\rho2\|\Gamma-Z^q+U^q\|_F^2,\nonumber\\
 \widetilde Z^{q+1}&=\operatorname{prox}_{(\lambda/\rho)P}(\Gamma^{q+1}+U^q),\nonumber\\
 Z^{q+1}&=\operatorname{Snap}_{\tau}(\widetilde Z^{q+1}),\nonumber\\
 U^{q+1}&=U^q+\Gamma^{q+1}-Z^{q+1},
 \label{eq:admm-updates}
\end{align}
where $P(Z)=\sum_{a<b}\|z_a-z_b\|_1$. The $\Gamma$ step uses 30 Adam steps per inner iteration. For each coordinate of $W=\Gamma^{q+1}+U^q$, sort its $K$ entries increasingly, obtaining $w_{(1)},\ldots,w_{(K)}$. The exact pre-snap proximal update is the nondecreasing isotonic projection of
\[
 \left(w_{(i)}-\frac\lambda\rho(2i-K-1)\right)_{i=1}^K,
\]
followed by undoing the sorting permutation \citep{lin2019clustered}. Appendix states why this solves the coordinate subproblem. The subsequent row snapping averages near-equal rows with $\tau=10^{-3}$; it is a separate numerical operation, not part of the exact proximal solution.

Partition extraction uses six-decimal rounding and a $10^{-6}$
row tolerance. We run at most four cycles with validation-guided
model selection. Matching and selection tolerances, warm starts,
stopping rules, and fallback choices are fixed before test
evaluation and specified in Appendix. The change score is a
parameterization-dependent numerical diagnostic.

\subsection{Fixed-representation optimization}
For fixed $\theta$ and $\lambda\geq0$,
Eq.~\eqref{eq:fusion-objective} is convex for squared or Huber loss.
The pre-snapping auxiliary proximal update is exact.
Standard two-block ADMM results apply to an idealized
fixed-representation solver with exact subproblem minimization,
fixed $\rho>0$, and a primal--dual saddle point
\citep[Section 3.2]{boyd2011distributed};
%Appendix~\ref{app:theory} gives the details.

The implementation combines finite Adam updates, snapping,
representation adaptation, and validation-selected grouped refits.
We assess its numerical behavior through inner-solver diagnostics
and full-pipeline sensitivity experiments.
Convergence of the full alternating procedure remains open.

\section{Experiments}
\label{sec:experiments}

\subsection{Design, baselines, and evaluation}
We evaluate the locked NSCF implementation on exact treatment sharing and departures from it. Covariates are independent $\operatorname{Unif}[-1,1]$ variables with $p=10$, treatments are assigned uniformly and independently of $X$, and
\begin{equation}
 Y=m_s(X)+\phi_s(X)^\top\{\eta_{g(A)}+\epsilon u_A\}+e.
 \label{eq:dgp}
\end{equation}
Here $e\sim N(0,0.5^2)$, $\eta_{g(A)}$ is a group prototype, and $u_a$ has independent standard-normal coordinates. The scalar $\epsilon$ controls within-prototype heterogeneity; $\epsilon=0$ gives exact sharing. The perturbations are fixed across training, validation, and test splits within a replication. The separation parameter multiplies the prototypes, not the perturbations; it is a coefficient-scale parameter rather than a minimum functional distance.

The contrast maps range from linear to nonlinear-easy and nonlinear-hard, with polynomial, trigonometric, and interaction terms. These grouped designs test the structural premise, not arbitrary response surfaces.

The primary comparison uses $K=12$, two groups, separation 1.5, and training/validation/test sizes $1000/500/10000$. Additional families use validation size $n/2$ and 10,000 test samples (Figures~\ref{fig:g-sample}--\ref{fig:g-groups}). Exact-sharing comparisons use 50 paired seeds (20 previously inspected plus 30 additional), except $n=400$, which retains 20. Methods share data and validation-only tuning; pooled summaries are not untouched confirmatory tests.

Comparators are a strongly trained unfused Neural Q network, ridge, lasso \citep{tibshirani1996lasso}, elastic net \citep{zou2005elasticnet}, random forest (RF) \citep{breiman2001random}, RBF support vector regression (SVM) \citep{drucker1997support}, and SCAF \citep{ma2022many}. NSCF denotes the locked discovery output described above. SCAF uses observed covariates with an intercept, not the oracle nonlinear basis. Classical fits use training data only and validation factual MSE for tuning; SCAF uses the same arrays.

We did not include an empirical comparison with \citet{zhu2025doubly}, as we were unable to locate a public implementation at the time of our experiments. We therefore make no claim of empirical superiority over their method.

% Declare the first two full-width figures ahead of their discussion so that
% the float queue does not collect three figures on one later page.
\begin{figure*}[!t]
 \centering\includegraphics[width=\textwidth]{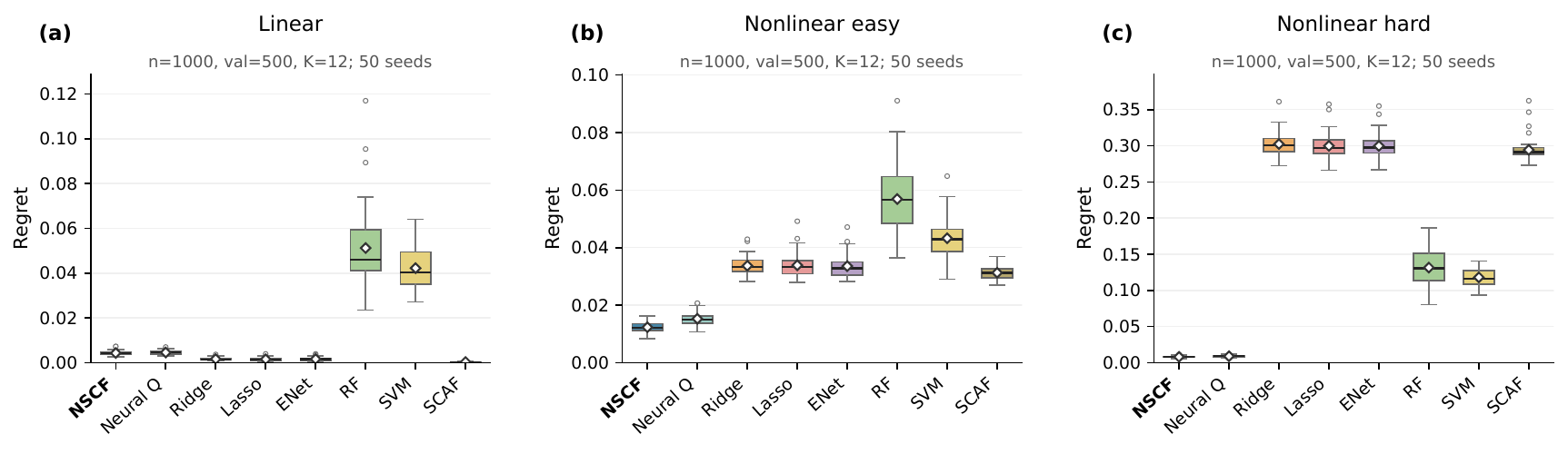}
 \caption{Exact sharing: eight methods and 50 paired seeds per scenario, $K=12$, two groups, training/validation sizes $1000/500$, separation 1.5. Diamonds denote means, box lines medians, whiskers 1.5 IQR, and circles observations beyond the whiskers; no seeds are excluded. Vertical scales differ across panels.}
 \label{fig:regret-scenarios}
\end{figure*}
\begin{figure*}[!t]
 \centering\includegraphics[width=\textwidth]{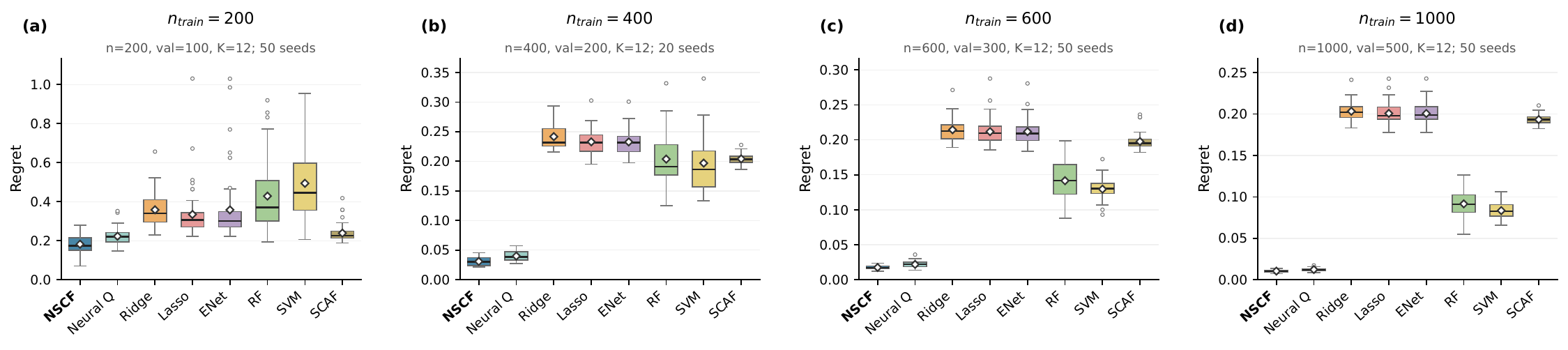}
 \caption{Sample-size comparison in nonlinear-hard: $K=12$, two groups, separation 1.0, validation size $n/2$. All eight methods are shown. The $n=200,600,1000$ panels use 50 paired seeds; $n=400$ uses 20. Vertical scales differ, with all outliers retained. Boxplot conventions follow Figure~\ref{fig:regret-scenarios}.}
 \label{fig:g-sample}
\end{figure*}

Neural Q shares NSCF's architecture and supplies its discovery anchor. Continued unfused training uses a prespecified per-replication update cap and validation early stopping. The calibration-based cap does not equalize total computation: NSCF additionally performs fusion and grouped refits.

The primary metric is test regret, computed from the noiseless true treatment means:
\[
 \widehat R(\widehat d)=\frac1{n_{\rm test}}\sum_i\left\{\max_a Q(X_i,a)-Q(X_i,\widehat d(X_i))\right\}.
\]
We also report adjusted Rand index (ARI), selected group count, and exact partition recovery. The units for uncertainty are simulation seeds (50 or 20 as specified), not the 10,000 test observations. Intervals are pointwise 95\% paired-$t$ intervals for seed-level regret differences, not simultaneous intervals. Value and regret use the same test outcomes and do not constitute separate evidence of improvement.

\subsection{Exact-sharing policy performance}
Figure~\ref{fig:regret-scenarios} shows structure-dependent benefits. In nonlinear-easy, NSCF mean regret is 0.01234 versus 0.01529 for Neural Q, a 19.3\% reduction with 49/50 paired wins. Nonlinear-hard gives 0.00808 versus 0.00904, a 10.7\% reduction with 45/50 wins. Both paired intervals exclude zero. %(Table~\ref{tab:g-exact}).

In the linear setting, NSCF and Neural Q are close (0.00434 versus 0.00458), while SCAF is substantially better (0.00033). Comparisons with linear fusion support nonlinear modeling in the nonlinear scenarios, but do not isolate representation learning's contribution to \emph{group discovery} from that of the final predictor.

\subsection{Sample size, group complexity, and visualization}
\begin{figure*}[!t]
 \centering\includegraphics[width=\textwidth]{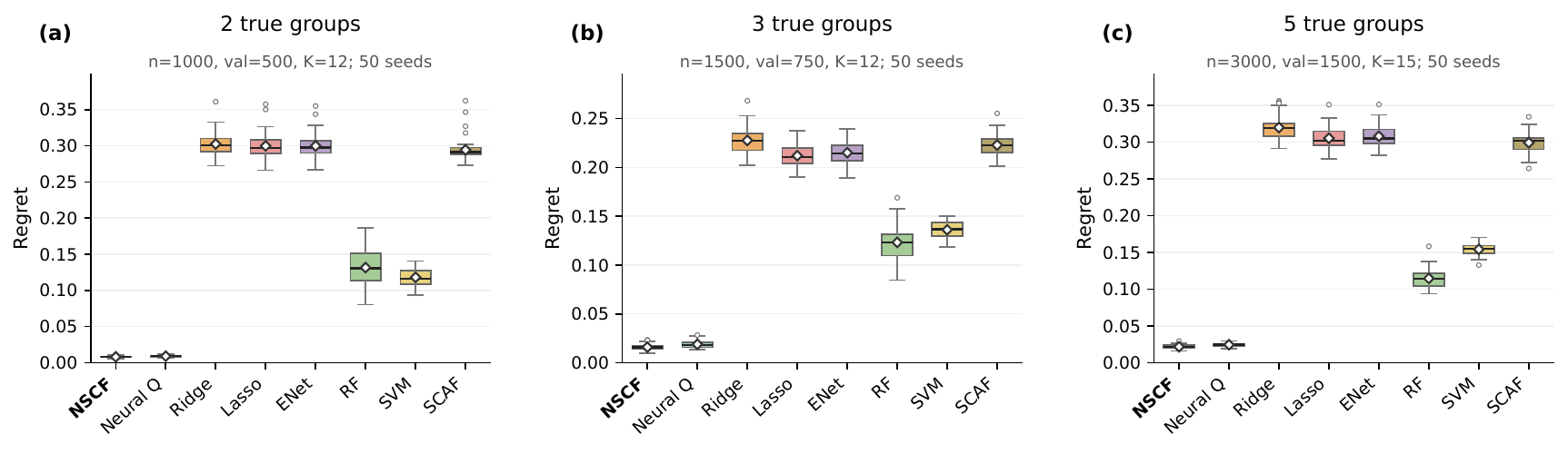}
 \caption{Treatment-group complexity in nonlinear-hard, separation 1.5: $(G^*,K,n)=(2,12,1000)$, $(3,12,1500)$, and $(5,15,3000)$. Each panel includes eight methods and 50 paired seeds. Sample size, architecture and sometimes $K$ vary, so this is not a one-factor group-count ablation.}
 \label{fig:g-groups}
\end{figure*}
In Figure~\ref{fig:g-sample}, NSCF mean regret decreases from 0.18214 at $n=200$ to 0.01042 at $n=1000$, versus 0.22278 to 0.01224 for Neural Q. All four sample-size comparisons favor NSCF in paired intervals. Exact recovery improves from 11/50 seeds at $n=200$ to 19/20 at $n=400$ and 50/50 at larger sizes.

For three and five groups (Figure~\ref{fig:g-groups}), NSCF reduces mean regret relative to Neural Q by 15.4\% and 11.2\%, with 49/50 and 46/50 wins.

\begin{figure*}[!t]
 \centering\includegraphics[width=\textwidth]{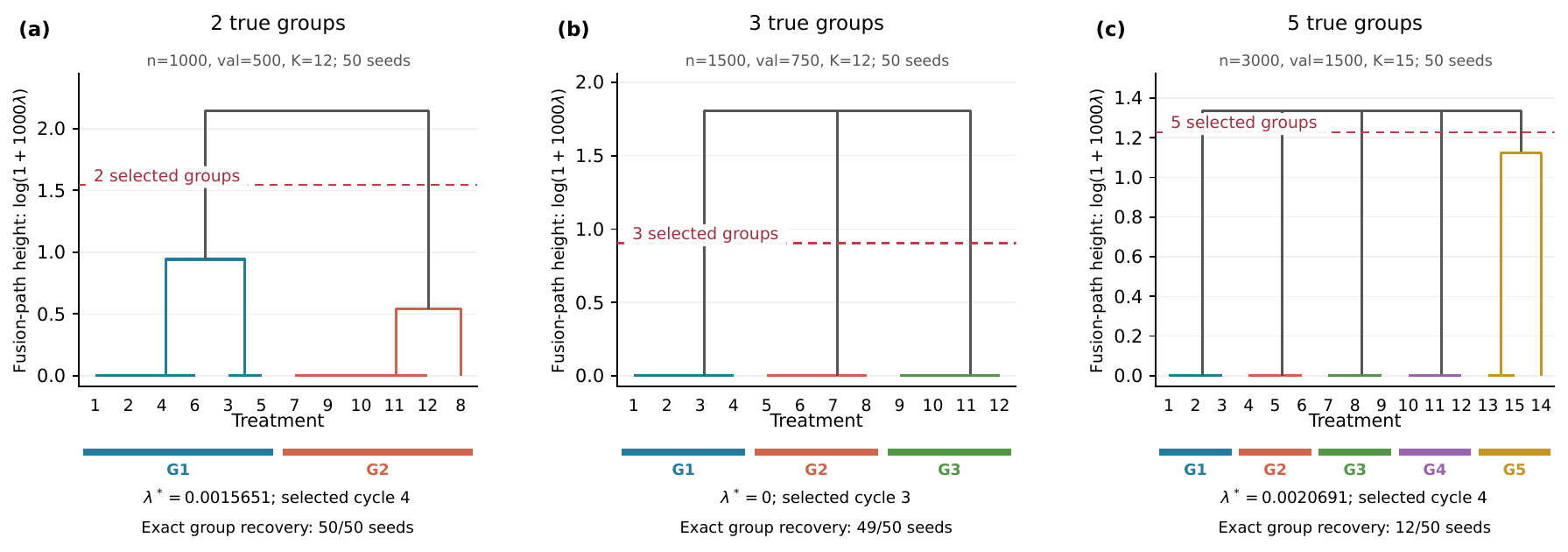}
 \caption{Recorded fusion paths for fixed seed 200, not selected by performance. Branch colors and bottom bars show selected groups. The dashed display cut lies between successive merge heights and reproduces the saved partition; it is not the selected $\lambda^*$, which is printed below each panel. Heights are $\log(1+1000\lambda)$. Exact recovery is 50/50, 49/50, and 12/50; the five-group example is not typical of recovery frequency.}
 \label{fig:simulation-dendrograms}
\end{figure*}
The example paths are nested, although general fusion paths need not be hierarchical. Zero-height merges can reflect inherited groups: later-cycle $\lambda=0$ need not represent an independently trained unfused model.

Five-group recovery is less complete: mean ARI is 0.888, mean group count 6.46, and exact recovery 12/50. Good policy performance does not imply exact recovery; the partition represents predictive sharing.

\subsection{Mild heterogeneity and the boundary of hard sharing}
\begin{figure*}[!t]
 \centering\includegraphics[width=\textwidth]{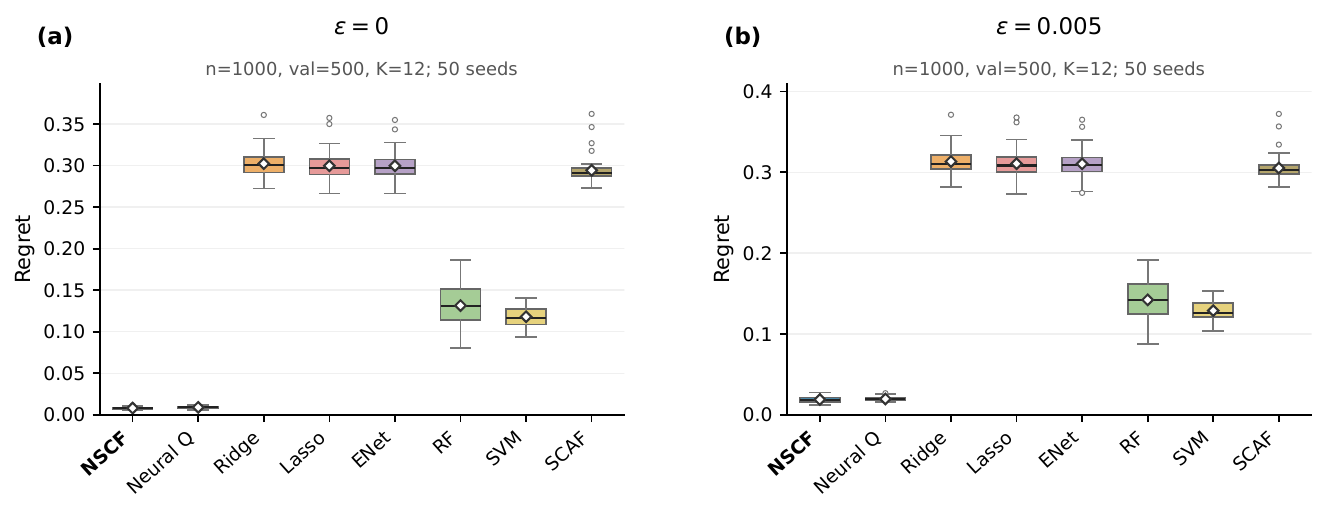}
 \caption{Exact and very mild sharing: nonlinear-hard, $\epsilon=0$ and 0.005, $K=12$, two prototypes, separation 1.5, and a $1000/500/10000$ split. All eight methods and 50 paired seeds are retained. The exploratory 0.005 comparison gives a small mean advantage, with an interval including zero. }%Appendix~\ref{app:boundary} retains the complete range through $\epsilon=0.20$.}
 \label{fig:mild}
\end{figure*}

We perturb treatment prototypes in Eq.~\eqref{eq:dgp}, sharing covariates, assignments, noise streams, and perturbation directions across $\epsilon$ within each seed. Candidate grids and validation rules remain fixed; tuning does not use the true perturbation size.

At $\epsilon=0.005$, NSCF mean regret is 0.01894 versus 0.01960 for Neural Q, a 3.4\% reduction (Figure~\ref{fig:mild}); the paired interval $[-0.00159,0.00027]$ does not establish superiority. The methods are approximately tied at $\epsilon=0.01$, and larger perturbations favor Neural Q.

The .005/.01 levels were added after inspecting the wider grid and expanded to 50 seeds without changing selection rules; larger perturbations retain 20. These pooled comparisons are exploratory. Positive $\epsilon$ creates decision-relevant within-group differences, so regret takes priority over prototype ARI; the transition is not a universal threshold.

\subsection{Role of alternating updates and grouped refitting}
Development-cohort ablations favor updating the representation over frozen features in both nonlinear scenarios. Additional cycles improve nonlinear-easy prediction, but one cycle captures most of the nonlinear-hard benefit. One cycle includes grouped refitting, not a no-refit control. Common-partition restarts assess refit duration and head-only versus full-network updates; a longer restart need not outperform NSCF's selected predictor. These seeds are not pooled with the locked benchmark.

\paragraph{Penalty sensitivity.}
A fixed-representation comparison uses 20 paired development seeds per scenario. With frozen neural features, $\ell_1$ and group-$\ell_2$ select the same grouped-refit partition in 59/60 cases. Penalized predictions differ, but post-grouping refits perform similarly.

This supports empirical penalty robustness in these exact-sharing settings, not universal $\ell_1$ superiority. Group-$\ell_2$ candidate paths are faster in this implementation; the comparison does not replace the full alternating algorithm with group-$\ell_2$.

\subsection{Numerical diagnostics}
In 60 locked exact-sharing runs, selected models show small relative primal residuals and limited late-stage objective variation. Outer cycles retain partitions while refining predictions. These numerical diagnostics are not convergence guarantees or evidence of insensitivity to solver settings.

Numerical checks on ten paired nonlinear-hard development seeds separately vary the ADMM budget, $\rho$, and snapping threshold. All configurations recover the same two-group partition. Mean regret ranges from 0.00819 to 0.00852, versus 0.00844 by default, although individual predictions differ. All configurations and seeds are retained.

\FloatBarrier
\begin{figure*}[!t]
 \centering\includegraphics[width=\textwidth]{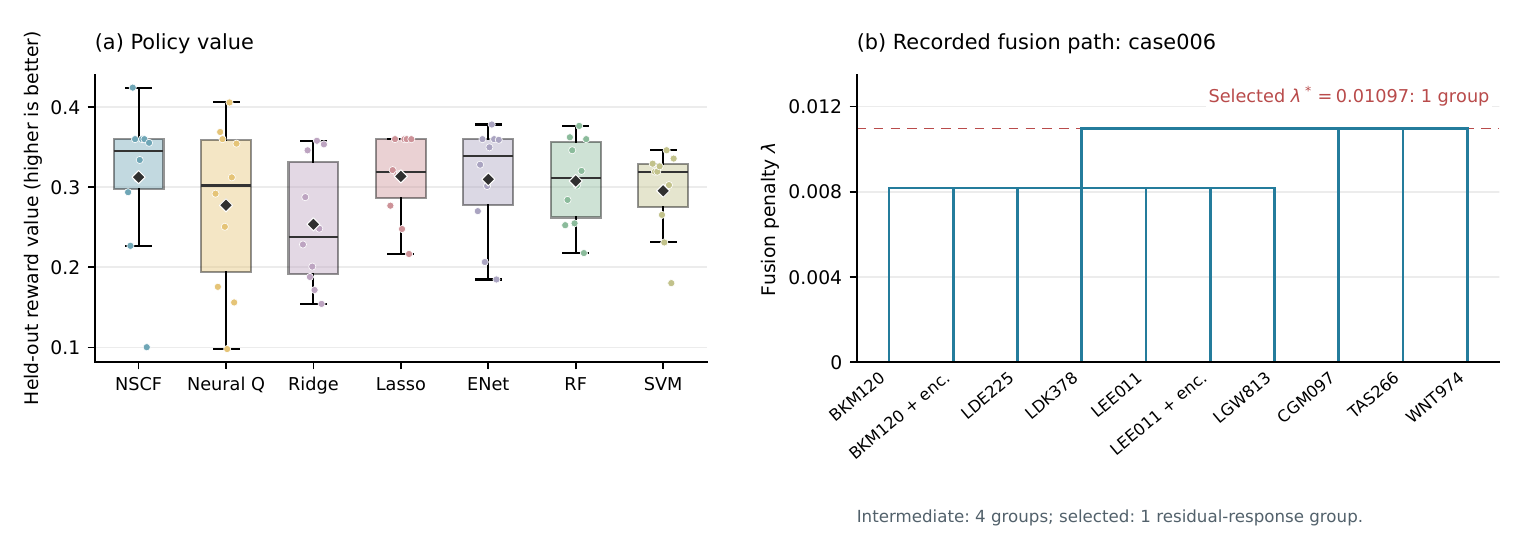}
 \caption{Melanoma PDX application. (a) Held-out reward ($-\texttt{RespScaled}$; higher is better): ten complete five-fold repeats of the same 32 models; diamonds denote means. (b) Recorded fusion path for fixed case006, selected cycle 4. Heights are evaluated $\lambda$ values, not drug distances; tied merges imply no internal order. The dashed line marks the selected single residual-response group, not the intermediate four groups. Treatment means remain unfused; enc. denotes encorafenib. This is not a cross-fold consensus or drug-mechanism hierarchy.}
 \label{fig:pdx-value}
\end{figure*}
\section{Application to melanoma xenografts}
\label{sec:realdata}
We analyze a complete panel of 32 cutaneous melanoma patient-derived xenograft (PDX) models under ten treatments from \citet{gao2015high}. Ten repeated five-fold splits keep each PDX model's treatment observations together. Feature processing and treatment means use training models only; model selection uses validation MSE. Each split reserves six models for validation and six or seven for testing. %Appendix~\ref{app:pdx} details the protocol and preprocessing sensitivity analysis.

The processed data provide 32,753 candidate genomic features. A training-only lasso-path screen targets treatment-mean residuals and retains at most 100 active features. All methods use the same splits and screened inputs within each preprocessing analysis. Repeats assess split sensitivity; they do not increase the biological sample size.

The outcome, \texttt{RespScaled}, is a positive scaling of tumor-volume response, so \emph{smaller} is better. Models predict this response, recommend its minimum, and are evaluated with reward $-\texttt{RespScaled}$. Because each held-out model has all ten treatment outcomes, value is the mean observed reward under the recommended treatment. Our PDX implementation retains unpenalized training-estimated treatment means and fuses the residual-response component. With one residual-response group, predictions can differ through the unfused means, but the recommended treatment is constant across covariates.

Best constant recommends the treatment with the smallest training mean to every test model. Shared ridge adds a common covariate predictor, changing prediction error but not treatment ranking. These training-only diagnostic comparators, reported in Appendix, test whether individualization adds value beyond a cohort-wide recommendation.

NSCF attains mean held-out value 0.3124, compared with 0.2773 for strong Neural Q, 0.3134 for lasso, 0.3097 for elastic net, 0.3076 for RF, and 0.2955 for SVM (Figure~\ref{fig:pdx-value}). Its paired advantage over Neural Q averages 0.0351: seven repeats are positive, one tied, and two negative. Against lasso, NSCF wins four repeats, ties two, and loses four; their mean values are close. The constant strategy obtains 0.3328, so the cohort does not establish additional policy-value benefit over a constant recommendation.

NSCF selects one residual-response group in 33/50 folds. Across 320 model-by-repeat recommendations, 88.1\% select LEE011 plus encorafenib and 10.0\% select BKM120 plus encorafenib. This recommendation is consistent with activity reported in the same PDX screen \citep{gao2015high}, but does not constitute independent validation. The grouping and constant-policy comparison suggest that treatment-level differences dominate decisions in this panel. %A post-hoc RNA transformation sensitivity analysis is reported in Appendix~\ref{app:pdx}.

Figure~\ref{fig:pdx-value}(b) illustrates a recorded $10\to4\to1$ fusion path. The four-group configuration is intermediate; validation selects one group. This single-split example describes residual predictive sharing, not pharmacological equivalence or a population treatment hierarchy. Clinical subgroup claims require external validation.

\FloatBarrier
\section{Discussion}
\label{sec:discussion}
NSCF combines outcome-trained nonlinear features, treatment fusion, and grouped refitting. It improves on a strongly trained unfused neural reference in the nonlinear exact-sharing simulations; ablations support adapting the representation under grouping constraints. In the PDX application, it also attains higher average policy value than the neural reference, while the constant strategy remains competitive and has the highest mean. Our contribution is to integrate explicit treatment grouping with
neural representation adaptation in a validation-guided procedure
for treatment-rule learning.

Sharing is most useful when treatment effects exhibit common structure. Linear adaptive fusion remains preferable in the linear calibration case, and the perturbation study identifies settings where hard sharing loses decision-relevant differences. Validation factual MSE does not automatically detect this boundary. The PDX analysis further distinguishes residual-response sharing from absolute treatment efficacy: shared labels do not establish pharmacological equivalence or individualization gains.

Our empirical evaluation focuses on controlled simulations and a
small preclinical cohort. Generalization to observational settings
with confounding and broader model misspecification requires further
investigation. Numerical diagnostics assess the implemented procedure,
while the convergence analysis described above concerns an idealized
fixed-representation solver. The fixed-feature penalty comparison
shows similar post-refit performance in the tested settings.
Future work includes full-pipeline group-$\ell_2$ comparisons,
representation-scale sensitivity, and extensions accommodating
within-group heterogeneity, evaluated through policy performance
and group recovery.

% \section*{AI use statement}
% \input{sections/ai_statement}

\bibliographystyle{apalike}
\bibliography{references}

% \clearpage
% \appendix
% \section*{Supplementary Materials}
% \setcounter{figure}{0}
% \setcounter{table}{0}
% \renewcommand{\thefigure}{S\arabic{figure}}
% \renewcommand{\thetable}{S\arabic{table}}
% \input{sections/appendix}
% \input{sections/pdx_appendix}
% \input{sections/penalty_appendix}
\end{document}